\documentclass[11pt]{article}

\usepackage{acl}

\usepackage{times}
\usepackage{latexsym}

\usepackage[T1]{fontenc}

\usepackage[utf8]{inputenc}

\usepackage{microtype}

\usepackage{inconsolata}

\usepackage{graphicx}
\newcommand{\model}{\textit{SBS}}
\usepackage{booktabs}
\usepackage[ruled,vlined]{algorithm2e}
\usepackage{amsmath}
\usepackage{multirow}
\usepackage{array}
\usepackage{xcolor}
\usepackage[table]{xcolor}
\usepackage[dvipsnames]{xcolor}
\usepackage{pifont}
\definecolor{cGrey}{HTML}{eff3fa}

\usepackage{amsfonts} 
\usepackage{amssymb}  
\usepackage{cleveref}
\usepackage{makecell}   
\usepackage{enumitem}
\AtEndPreamble{
    \crefname{section}{Sec.}{Secs.}
    \Crefname{section}{Section}{Sections}
    \Crefname{table}{Table}{Tables}
    \crefname{table}{Tab.}{Tabs.}
    \Crefname{figure}{Figure}{Figures}
}

\title{Seeing Before Synthesizing: VLM-Guided Transition Event Discovery \\ for Weakly-Supervised Dense Video Captioning}

\author{\textbf{Ye-Chan Kim}\quad
        \textbf{Seung hee Choi}\quad
        \textbf{SeungJu Cha}\quad
        \textbf{Si-Woo Kim} \\
        \textbf{Hwiseon Kim}\quad
        \textbf{Hyungee Kim}\quad
        \textbf{Dong-Jin Kim}$^{\dagger}$ \\
        Hanyang University, South Korea \\
        {\footnotesize\texttt{\{dpcksdl78, ermitaju1, sju9020, boreng0817, hwiseon9151, khjiiii2002, djdkim\}@hanyang.ac.kr}}}

\begin{document}
\maketitle

\begin{abstract}
Weakly-Supervised Dense Video Captioning aims to localize and describe multiple events in untrimmed videos given only an ordered set of event-level captions per video.
Recent work synthesizes auxiliary transition captions via LLM to provide additional vision-language alignment, but these captions lack visual grounding and are rigidly assigned to every inter-event gap at a fixed location and duration.
To address these, we propose Seeing Before Synthesizing ($\model$), a framework that adaptively provides visually grounded linguistic guidance only where warranted. 
Leveraging a VLM, we generate frame-level narratives for the inter-event gaps and detect transitions from the semantic variation across them.
For identified transitions, we then refine inter-event temporal masks by blending the temporal midpoint with the semantic change point and selecting the width that maximizes vision-language alignment.
Experiments on ActivityNet Captions and YouCook2 demonstrate state-of-the-art performance in both captioning and localization.
\end{abstract}
\section{Introduction}
\begin{figure}[t]
\begin{center}
\includegraphics[width=\columnwidth]{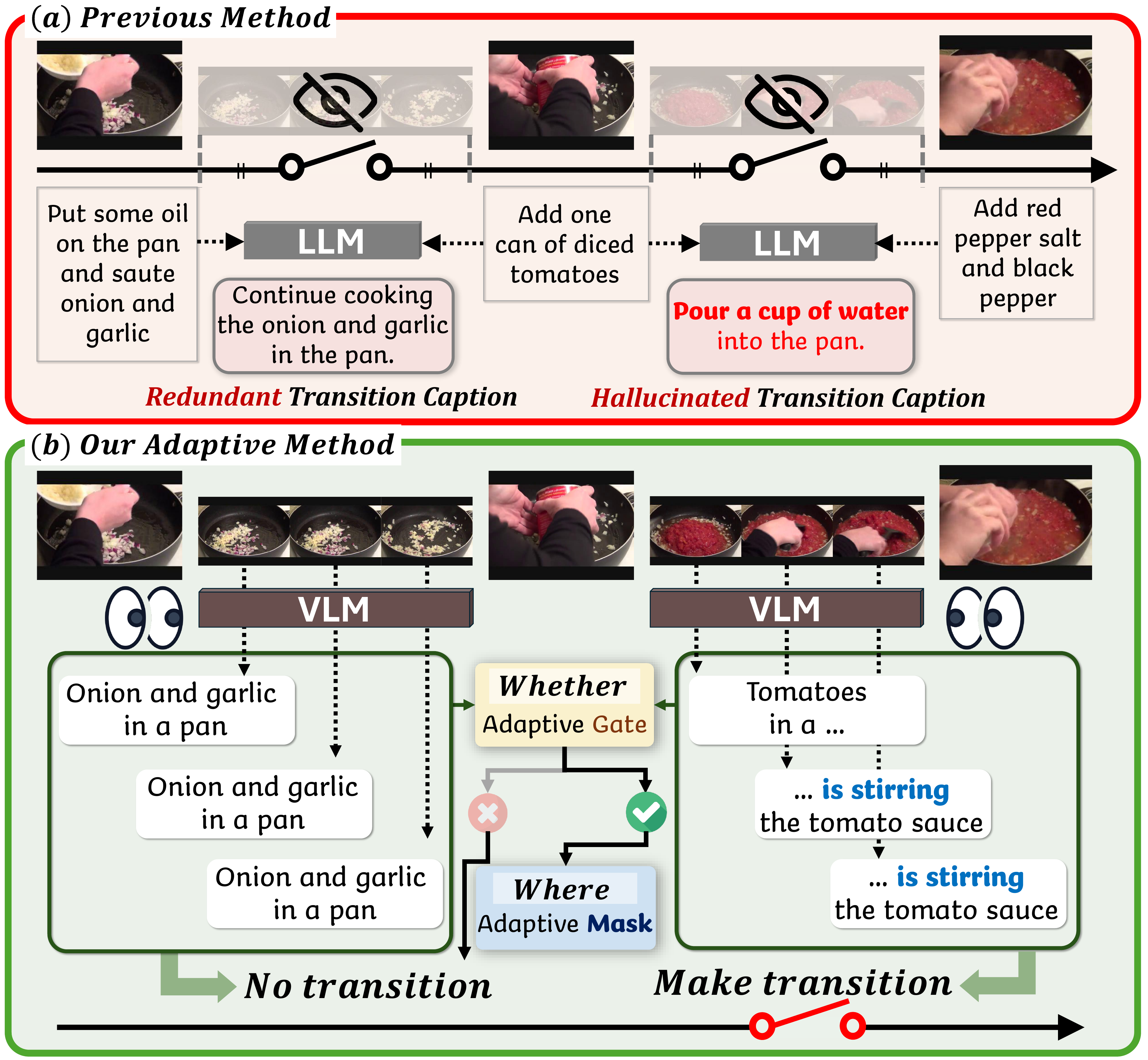}
\end{center}
\caption{(a) Prior work~\cite{sail} synthesizes captions from GT text alone, producing hallucinated descriptions (e.g., ``pour a cup of water'') and rigidly bridging all gaps. (b) $\model$ generates visually grounded captions via VLM and selectively bridges gaps only when a genuine semantic transition is detected.}
\label{fig:intro_fig}
\end{figure}
Dense Video Captioning (DVC)~\cite{cm2,ddvc,e2dvc} extends standard video captioning~\cite{wang2018video, seo2022video, zhao2023video} by localizing and describing multiple temporal events in long, untrimmed videos. Conventional fully supervised DVC methods~\cite{starc,stay} typically rely on dense annotations, where each event is paired with both temporal boundaries and a natural-language description. 
However, obtaining such fine-grained annotations is labor-intensive and difficult to scale, especially for real-world videos.

To alleviate this dependency, Weakly-Supervised Dense Video Captioning (WSDVC)~\cite{WSDEC,ilcacm,sail} has emerged.
It learns event localization and captioning from videos paired with temporally ordered descriptions, eliminating the need for start and end timestamps.
In this setting, each training video provides the sequence of event-level captions, but the temporal boundaries of the events remain unannotated.
Therefore, effectively aligning language supervision with the underlying visual content is crucial for learning fine-grained event localization.

To better support such Vision-Language (VL) alignment, recent studies have explored Large Language Model (LLM)-generated captions as auxiliary supervision when annotations are sparse or noisy~\cite{dibs,howtocaption,laclip}.
Since VL alignment serves as a crucial learning cue in WSDVC, SAIL~\cite{sail}, an early attempt to apply this idea, uses an LLM to synthesize transition captions for each inter-event gap—the interval between two neighboring predicted event centers.
However, such approaches are not grounded in the visual content and resort to rigid, heuristic rules. 
As a result, SAIL has no basis for the two decisions most essential to providing useful transition supervision in WSDVC: \textit{whether} an inter-event gap actually contains a transition worth describing, and \textit{where} within that gap the transition occurs.
This leads to two key limitations.


\textbf{First}, regarding \textit{whether} to 
introduce a caption, the existing method~\cite{sail} blindly assumes that every inter-event gap {must contain a transitional event}.
However, this rigid assumption ignores the diverse nature of real videos, where some transitions are already sufficiently covered by adjacent Ground-Truth (GT) captions.
As a result, assigning auxiliary transition captions to every gap regardless of the video content adds redundant captions to already well-described regions, introducing noise rather than useful cues.
Moreover, since each transition caption is generated solely from the surrounding GT event descriptions, it is prone to hallucinating content that misrepresents the video (\Cref{fig:intro_fig}).
\textbf{Second}, even when a transition does exist, each synthesized transition caption is misaligned with its visual region by a {fixed rule}.
In particular, prior work~\cite{sail} assumes the transition always lies at the midpoint between two neighboring events and spans a fixed duration, without adapting to the observed transition pattern.
This can harm model training when the actual transitional event is displaced from the midpoint or covers a different duration. 

To move beyond such rigid assumptions, we aim to provide transition cues that are selective and visually grounded, adaptive to each video's content.
Building on this goal, we propose {\textit{\textbf{S}eeing \textbf{B}efore \textbf{S}ynthesizing}} (\model), a framework that employs a Vision-Language Model (VLM)~\cite{li2023blip2}'s eye into both decisions—\textit{whether} to introduce a transition and \textit{where} to place it.
It selectively synthesizes a transitional event guided by contextual flow, only when an informative event occurs, and places it according to the video content.

{To realize this, the key question is determining \textit{whether} an informative event exists within each inter-event gap.}
Inspired by the observation in cognitive science that humans segment continuous activity into discrete events at points of substantial perceptual change~\cite{tversky2013event}, we utilize the \textbf{semantic change} across frames as a proxy for an unannotated transition.
A naive way to capture such a semantic change in video is to track fluctuations in low-level visual features directly.
However, these signals are notoriously susceptible to non-semantic noise such as camera motion and lighting changes~\cite{smeaton2010video,schiappa2022robustness}, which can easily be mistaken for genuine event transitions. 
We therefore repurpose the VLM as a transition-search tool, rather than a mere caption generator. 
Specifically, we exploit it to translate each frame into a semantically abstract linguistic description.
The key advantage is that linguistic descriptions provide a more semantically abstract signal than raw visual features~\cite{ye2025improving}, making transitions easily discernible, even when the visual appearance remains similar.

Concretely, we feed the semantic variation between caption embeddings of consecutive frames within each inter-event gap into an adaptive gate, whose threshold is determined by the local variation statistics of that gap.  
If the signal surpasses the threshold—indicating a salient change—the gate treats the gap as a genuine transition and opens to activate inter-event supervision, closing otherwise.

Once the adaptive gate identifies a meaningful transition, the remaining challenge is \textit{where} to align the transition caption within the visual region.
{Prior work addresses this with a fixed temporal midpoint between neighboring event centers, which remains blind to the actual video content.}
{Instead, we identify a content-adaptive transition center by leveraging the semantic change point that exhibits the greatest variation.}
{We interpolate between the semantic change point and the midpoint, anchoring the center near 
{the change point}
while keeping the transition event
from collapsing onto either of its neighboring event centers.}
{In addition, we select the width that best aligns the visual region with the transition caption.}
The resulting temporal span defines the visual region for the transition caption, supplying additional alignment to the model.

Our contributions are as follows:
\begin{itemize}
    \item We reformulate transition augmentation in WSDVC from text-only synthesis to visually grounded transition event discovery.
    \item We propose $\model$, which repurposes a VLM as a transition-search tool to adaptively gate transition cues and localize transition regions through a semantic change point.   
    \item We validate our method on ActivityNet and YouCook2, achieving state-of-the-art results in both
    the captioning and localization tasks.
\end{itemize}

\section{Related Work}
\noindent{\textbf{Weakly-Supervised Dense Video Captioning.}}
In WSDVC, early approaches~\cite{WSDEC,EC-SL} employ a cycle-consistency framework that localizes temporal segments from captions and reconstructs the captions from those segments.
Recently, distinct from existing cycle-consistency frameworks, ILCACM~\cite{ilcacm} proposed a method that employs Gaussian masks to construct event-specific visual features, implicitly learning localization and captioning through a reconstruction objective.
This reconstruction objective implicitly drives the simultaneous learning of both localization and captioning without relying on a cycle system. 
Building upon this, SAIL~\cite{sail} extended the ILCACM framework by introducing the concept of ``transitional events''. 
This approach utilizes an LLM to synthesize plausible captions for the intervals between given events based on their adjacent captions. 
By providing such auxiliary language information, SAIL enables the model to delineate more fine-grained event boundaries.
However, the existing method only offers a naive application of inter-events, leaving the question of how to effectively leverage them while accounting for video characteristics largely unexplored.

\noindent\textbf{{LLM-generated captions for Vision-Language tasks.}}
In recent years, the remarkable success of LLMs across various language tasks~\cite{gpt4, touvron2023llama} has demonstrated their exceptional zero-shot capabilities and common sense inference~\cite{cot}.
This success has promoted extensive research into integrating common sense knowledge into vision-language tasks~\cite{laclip,lskd}.
Notably, HowToCaption~\cite{howtocaption} addresses the fact that ASR subtitles only loosely correspond to the visual content by prompting an LLM to enrich these noisy narrations into human-style video captions.
Similarly, DIBS~\cite{dibs} targets the absence of dense annotations in unlabeled videos by exploiting diverse LLMs to generate rich, event-centric caption candidates.
However, since such enriched pseudo-labels often contain noisy or misaligned content, selectively leveraging this supplementary information remains underexplored.

\section{Proposed Method}
\begin{figure*}[t]
\begin{center}
\includegraphics[width=0.9\textwidth]{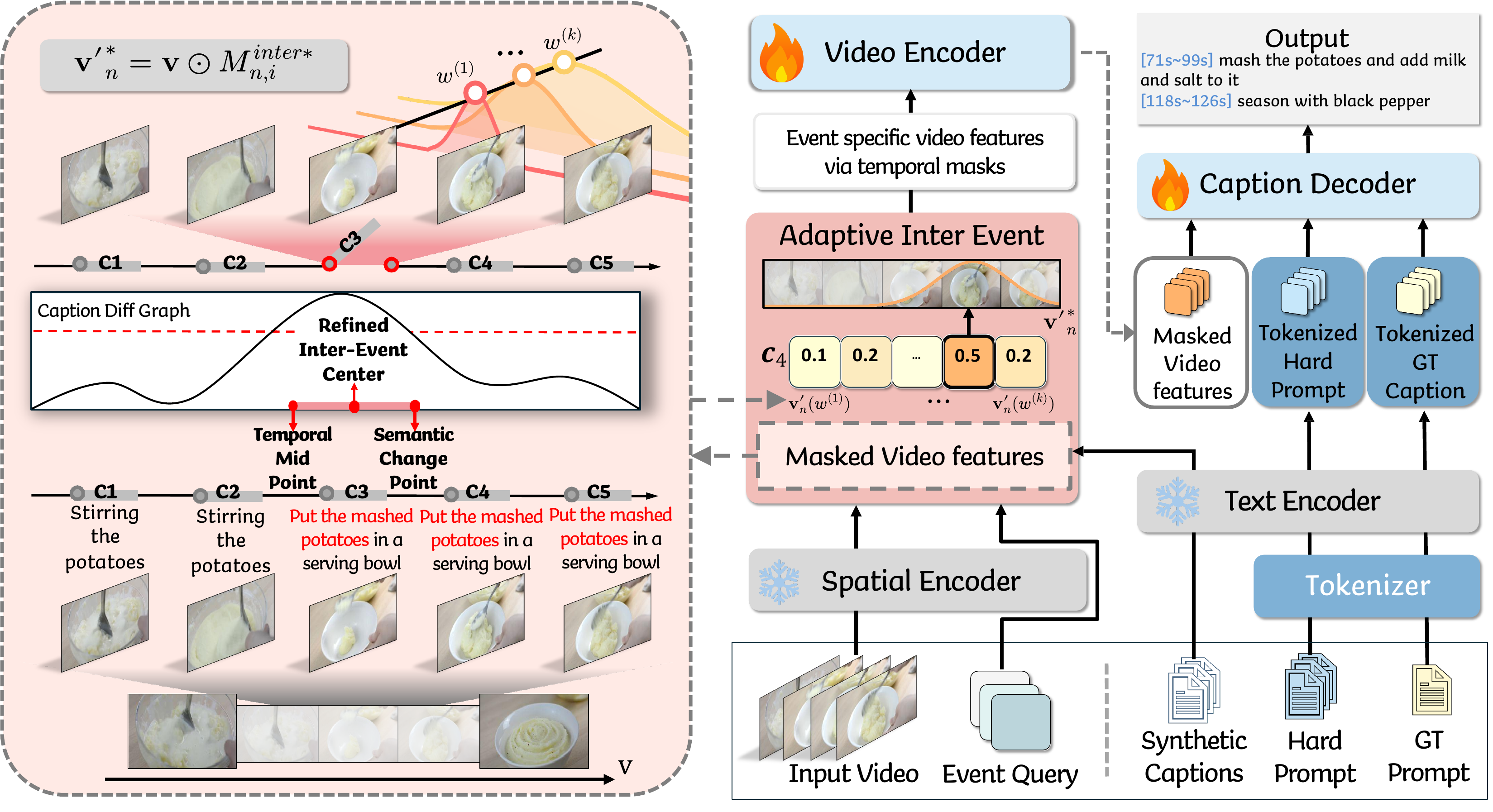}
\end{center}
\caption{$\model$ Pipeline.
Using VLM-generated frame captions as a visually grounded narrative flow, $\model$ decides \textit{whether} to introduce a transition in each gap via an adaptive gate and \textit{where} to place it via an adaptive mask, providing grounded auxiliary supervision.}
\label{fig:main}
\end{figure*}

Our objective is to effectively capture the genuine transitions to improve both captioning and localization performance in WSDVC. 
Formally, given a video $V$ containing $N_e$ distinct events, the goal is to generate a set of event timestamps and corresponding captions $(t_n^s, t_n^e, C_n)_{n=1}^{N_e}$, where $t_n^s$ and $t_n^e$ denote the start and end times of the $n$-th event, and $C_n$ represents the caption describing the $n$-th event.
Since temporal annotations are unavailable, the model learns to infer temporal event boundaries by aligning video frames with their corresponding textual descriptions. The overall architecture is shown in~\Cref{fig:main}.

\subsection{Preliminaries}
Following~\cite{ilcacm}, we use a differentiable Gaussian mask to represent each event region in the video.
To generate these masks, we employ a Transformer decoder taking video features with $N_v$ frames ${\mathbf{v}} = \{{v}_i\}_{i=1}^{N_v}$, extracted from the video $V$ using CLIP ViT-L/14~\cite{vit,clip}, and learnable event queries $\mathbf{q}_n$ to produce event-specific representations $\mathbf{o}_n$.
Based on $\mathbf{o}_n$, we predict its temporal center $c_n$ and width $w_n$ for each event:
$c_n = \textit{Sig}(\text{FC}_c (\mathbf{o}_n)){\in [0, 1]}, \quad w_n = \textit{Sig}(\text{FC}_w (\mathbf{o}_n)){\in [0, 1]}$.
Here, \textit{Sig}($\cdot$) denotes the sigmoid function, and $\text{FC}_c(\cdot)$, $\text{FC}_w(\cdot)$ are linear layers for predicting $n$-th event's $c_n$ and $w_n$.
We then construct Gaussian-based temporal masks to represent each event within the video:
\begin{equation}
    M_{n,i}^{evt} = \mathcal{G}(r_i;\, c_n,\, w_n) 
    = \exp\!\left(-\frac{(r_i - c_n)^2}{2(w_n / \tau_m)^2}\right),
\label{eq:mask_func}
\end{equation}
where $r_i$ $\in$ [0, 1] represents normalized temporal positions across the video $r_i=\frac{i-1}{N_v-1}, i\in\{1,\dots,N_v \}$, and $\tau_m$ is a hyperparameter controlling mask sharpness.

SAIL~\cite{sail} further constructs inter-event masks to align the captions of transition events with the visual representation, capturing the continuous narrative between events.
These masks are aligned with LLM-synthesized transition captions, which act as an indirect training signal.
For each interval, they construct a static inter-event mask centered at the midpoint of adjacent centers $c^{inter}_n = \frac{c_n + c_{n+1}}{2}$ with a predefined fixed width $w^{inter}$. 
These masks serve as a bridge between the $n$-th and $(n+1)$-th events, providing a consistent temporal prior for the inter-event regions.

\subsection{Adaptive Inter-Event}
Unlike prior work that exclusively relies on video-blind GT captions to construct transition events, we ground auxiliary supervision in the visual content, leveraging the semantic progression of frame-level captions to adaptively identify transition events.

\noindent\textbf{Narrative Generation.}
Given an input video of $N_v$ frames, we utilize BLIP-2~\cite{li2023blip2} to generate a frame-specific caption $\mathcal{C}_i$ for each frame, yielding a temporal sequence $\mathcal{C} = \{\mathcal{C}_1, \dots, \mathcal{C}_{N_v}\}$.
This sequence $\mathcal{C}$ serves as a dense narrative flow over the visual stream.
Building upon the frame-level narrative flow, we introduce Narrative-Aware Inter-Event Selection to determine whether a latent transition event exists within the temporal gaps between consecutive predicted events.

\noindent\textbf{Narrative-Aware Inter-Event Selection.}
We leverage the intuition~\cite{tversky2013event} that stable semantic content yields little variation across frames, whereas an underlying event transition triggers a pronounced narrative shift.
To capture this shift, we measure the semantic dissimilarity within each inter-event gap.
For the $n$-th and $(n+1)$-th consecutive predicted events,  we define the inter-event gap as the temporal span between their centers from $c_n$ to $c_{n+1}$.
We map a normalized temporal coordinate $x \in [0,1]$ to a discrete frame index using
$\kappa(x)=
\operatorname{clip}\left(
\left\lfloor x (N_v-1) \right\rfloor + 1,
1,
N_v
\right).$

Then, the frame indices of the $n$-th inter-event search interval are given by $b_n^s = \kappa(c_n), b_n^e = \kappa(c_{n+1})$.
Within this temporal span, we collect the corresponding sequence of frame-level text embeddings $\{\mathbf{z}_i\}_{i=b_n^s}^{b_{n}^e}$, where $\mathbf{z}_i$ denotes the text embedding of caption $\mathcal{C}_i$ encoded via a CLIP text encoder.
To quantify the semantic shift within the gap $[b_n^s,\dots, b_{n}^e]$, we compute the cosine dissimilarity $d_i$ between adjacent frame-level narratives:
 \begin{equation}
    d_i = 1 - \frac{\mathbf{z}_i \cdot \mathbf{z}_{i+1}}{\|\mathbf{z}_i\| \|\mathbf{z}_{i+1}\|},
\end{equation}
where $d_i$ represents the degree of semantic variation at a specific time step.
A high dissimilarity within the gap's sequence $\mathcal{D}_n = \{d_i\}_{i=b_n^s}^{b_n^e - 1}$ signals a potential transition.
We thus feed this signal into an adaptive gating mechanism that decides whether a transition event should be synthesized.


Specifically, the gate determines the existence of an inter-event transition through a threshold adapted to each gap.
We compute the mean $\mu({\mathcal{D}_n})$, and the standard deviation $\sigma({\mathcal{D}_n})$ within the $n$-th gap.
The adaptive threshold $\eta^{adap}_n$ is then formulated as: $\eta^{adap}_n = \mu({\mathcal{D}_n}) + \beta \cdot \sigma({\mathcal{D}_n})$, where $\beta$ is a hyperparameter controlling the sensitivity to semantic shifts, following~\cite{sali4vid}.

To obtain a soft confidence score for each candidate inter-event interval, we employ a sigmoid-based gate $g_n$ to estimate the probability of an inter-event's existence: 
\begin{equation}
 g_n = \textit{Sig}(\text{max}(\mathcal{D}_{n}) - \eta^{adap}_n). 
 \label{eqn:gate}
\end{equation}
The gate value $g_n \in [0, 1]$ serves as a confidence score for the existence of a transition within the $n$-th gap.
We open the gate when $g_n \geq 0.5$, otherwise the gate is closed, and no transitional event is injected into the gap.
We aim to apply transition supervision selectively, rather than uniformly across all gaps. 
Through this selective mechanism, a transition cue is injected only into gaps whose semantic variation sufficiently exceeds the threshold.

\noindent\textbf{Adaptive Inter-Event Masks.}
While the gating mechanism identifies the necessity of a transition caption, the remaining challenge is to align it with its corresponding visual region. 
To achieve this, we propose Adaptive Inter-Event Masks, which move beyond video-agnostic templates by dynamically estimating the temporal center and width for each transition based on the video's flow and content.

Prior work~\cite{sail} uniformly assigns a static midpoint $c^{inter}_n$ and a fixed width $w^{inter}$ to every interval between the $n$-th and $(n+1)$-th events.
This rule serves as a temporal regularizer that keeps the transition event from collapsing onto either of its neighboring event centers.
In contrast, we refine the transition event's center to account not only for this temporal prior but also for the semantic change point identified in the narrative flow.
Specifically, we define this semantic change point $p_n$ as the temporal position exhibiting the largest semantic dissimilarity within the $n$-th gap:
\begin{equation}
    p_n = \frac{i_n^* - 1}{N_v - 1}, \quad \text{where} \quad i_n^* = \underset{i \in \{b_n^s,\dots, b_n^e-1\}}{\arg\max} \, d_i.
\end{equation}
Based on $p_n$, we refine the center $c_n^{inter*}$ with a hyperparameter $\alpha \in [0, 1]$:
\begin{equation}
c_n^{inter*} = (1 - \alpha) \cdot c_n^{inter} + \alpha \cdot p_{n}.
\end{equation}

As a result, the mask center is anchored at a position that reflects both the temporal prior from the neighboring centers and the point where the actual semantic change occurs. 
We then map $c_n^{inter*}$ to its corresponding discrete frame index 
$j_n = \kappa_{\mathrm{}}(c_n^{inter*})$, and adopt VLM caption at frame $j_n$ as the linguistic description of the transition.

Next, we adaptively determine the optimal temporal width $w^{inter*}_n$. To account for varying event durations and identify the width that best matches the underlying content, we evaluate cross-modal alignment between the video content and the caption at the refined center across multiple candidate widths.
{For each candidate width from the predefined set $w^{(k)} \in \Omega = \{w^{(1)},\dots,w^{(K)}\}$,}
we generate a soft Gaussian mask 
$M_{n,i}^{inter}(w^{(k)})$ at the refined center $c_n^{inter*}$ and compute the masked video features 
$\mathbf{v}_n'(w^{(k)}) = \mathbf{v} \odot M_{n,i}^{inter}(w^{(k)})$.
We then obtain the average pooled representation 
$\bar{\mathbf{v}}_n'(w^{(k)})$
and select the optimal width that maximizes the cross-modal alignment 
with the inter-event's caption embedding $\mathbf{z}_{j_n}$, recording the resulting alignment score $s_n^*$:
\begin{equation}
\begin{aligned}
w^{inter*}_n &= \arg\max_{w^{(k)} \in \Omega} \, 
\text{cos}\!\left(\bar{\mathbf{v}}_n'(w^{(k)}),\; \mathbf{z}_{j_n}\right), \\
s_n^* &= \max_{w^{(k)} \in \Omega} \, 
\text{cos}\!\left(\bar{\mathbf{v}}_n'(w^{(k)}),\; \mathbf{z}_{j_n}\right),
\end{aligned}
\end{equation}
where $\mathbf{z}_{j_n}$ is the CLIP text embedding of the VLM caption generated at frame $j_n$.

The final adaptive inter-event mask $M_{n,i}^{inter*}$ is formulated by substituting the optimal center $c_n^{inter*}$ and width $w_n^{inter*}$ into~\Cref{eq:mask_func}:
\begin{equation}
    M_{n,i}^{inter*} = \mathcal{G}\!\left(r_i;\, c_n^{inter*},\, w_n^{inter*}\right).
\label{eq:adaptive_mask}
\end{equation}
The resulting inter-event visual representation is obtained as 
$\mathbf{v'}_n^{*} = \mathbf{v} \odot M_{n,i}^{inter*}$, 
which serves as the visual feature for the transition between the 
$n$-th and $(n+1)$-th events.
By adaptively estimating these temporal parameters, $\model$ provides a more realistic and context-aware transitional event, thereby facilitating precise alignment between visual dynamics and auxiliary linguistic signals.

\begin{table*}[t!]
\centering
\resizebox{0.98\linewidth}{!}{
\begin{tabular}{c|c|c|ccccc|ccc}
    \toprule[2pt]
    \multirow{2}{*}{\textbf{Setting}} & \multirow{2}{*}{\textbf{Model}} & \multirow{2}{*}{\textbf{Features}} & \multicolumn{5}{c|}{\textbf{Captioning}} & \multicolumn{3}{c}{\textbf{Localization}} \\
    \cmidrule(lr){4-8} \cmidrule(lr){9-11}
    & & & \textbf{SODA\_c} & \textbf{METEOR} & \textbf{CIDEr} & \textbf{ROUGE-L} & \textbf{BLEU-4} & \textbf{R@Avg} & \textbf{P@Avg} & \textbf{F1} \\
    \midrule

    \multirow{4}{*}{\begin{tabular}[c]{@{}c@{}}\textbf{Fully}\\ \textbf{Supervised}\end{tabular}} 
    
    & CM$^2$~\cite{cm2}    & CLIP & 6.18  & 8.55 & 33.01 & --    & 2.38 & 53.71 & 56.81 & 55.21 \\
    & E$^2$DVC~\cite{e2dvc} & CLIP & 6.13  & 8.57 & 33.63 & --    & 2.43 & 54.67 & 57.70 & 56.14 \\
    & CACMI~\cite{cacmi}   & CLIP & 6.39  & 8.68 & 33.80 & --    & 2.44 & 55.89 & 58.05 & 57.10 \\
     & ROS-DVC~\cite{stay}   & CLIP & 6.45  & 8.45 & 35.04 & --    & 2.36 & 55.35 & 55.65 & 55.50 \\
    
    \midrule 

    \multirow{7}{*}{\begin{tabular}[c]{@{}c@{}}\textbf{Weakly}\\ \textbf{Supervised}\end{tabular}} 
    & WSDEC~\cite{WSDEC}     & C3D  & --    & 6.30 & 18.77 & 12.55 & 1.27 & 29.57 & 59.33 & 39.18 \\
    & ECG~\cite{ecg}       & C3D  & --    & 7.06 & 14.25 & --    & 1.33 & --    & --    & --     \\
    & EC-SL~\cite{EC-SL}    & C3D  & --    & 7.49 & 21.21 & 13.02 & 1.33 & --    & --    & --     \\
    & PWS-DVC$^*$~\cite{pws-dvc} & C3D  & --    & 7.28 & 20.59 & 12.71 & 1.35 & 40.85 & 55.82 & 47.09 \\
    & ILCACM~\cite{ilcacm} & CLIP & 6.08  & 8.48 & 33.42 & 14.77 & 2.26 & 53.72 & 58.92 & 56.20 \\
    & SAIL~\cite{sail} & CLIP & 6.29  & 8.63 & 35.38 & 15.29 & {2.30} & 54.39 & 59.87 & 57.00 \\
    
    & \cellcolor{cGrey}\textbf{$\model$ (Ours)} & \cellcolor{cGrey}CLIP & \cellcolor{cGrey}\textbf{6.49} & \cellcolor{cGrey}\textbf{8.87} & \cellcolor{cGrey}\textbf{36.87} & \cellcolor{cGrey}\textbf{15.60} & \cellcolor{cGrey}\textbf{2.47} & \cellcolor{cGrey}\textbf{56.13} & \cellcolor{cGrey}\textbf{60.38} & \cellcolor{cGrey}\textbf{58.18} \\
    \bottomrule[2pt]
\end{tabular}
}
\caption{Comparison with state-of-the-art methods on ActivityNet validation set. $\model$ achieves state-of-the-art performance in both captioning and localization metrics. * denotes results reported in~\cite{sail}.}
\label{table:main}
\end{table*}
\begin{table*}[t!]
\centering
\resizebox{0.98\textwidth}{!}{
\begin{tabular}{c|c|>{\hspace{1.5em}}c<{\hspace{1.5em}}|ccccc|ccc}
    \toprule[2pt]
    \multirow{2}{*}[-1.0ex]{\centering\arraybackslash\textbf{Setting}} & \multirow{2}{*}[-1.0ex]{\centering\arraybackslash\textbf{Model}}
    & \multirow{2}{*}[-1.0ex]{\centering\arraybackslash\textbf{Features}}
    & \multicolumn{5}{c|}{\textbf{Captioning}} 
    & \multicolumn{3}{c}{\textbf{Localization}} \\
    \cmidrule(lr){4-8} \cmidrule(lr){9-11}
    & & 
    & \textbf{SODA\_c} & \textbf{METEOR} & \textbf{CIDEr} & \textbf{ROUGE-L} & \textbf{BLEU@N}
    & \textbf{R@Avg} & \textbf{P@Avg} & \textbf{F1} \\
    \midrule 
    
    \multirow{5}{*}{\centering\arraybackslash\textbf{\shortstack{Weakly\\Supervised}}} 
    & WSDEC$^*$~\cite{WSDEC} & C3D & 2.11 & 1.47 & 8.43 & -- & -- & -- & -- & -- \\
   

      & PWS-DVC$^*$~\cite{pws-dvc} & C3D & 3.14 & 2.48 & 9.81 & -- & -- & -- & -- & -- \\
    
    & ILCACM$^*$~\cite{ilcacm} & CLIP & {3.60} & {3.41} & {13.49} & {4.75} & {2.59} & {17.76} & {18.01} & {17.88} \\

    & SAIL~\cite{sail} & CLIP & {4.08} & {3.63} & {14.61} & {5.42} & {2.94} & {20.76} & {21.13} & {20.94} \\

    & \cellcolor{cGrey}\textbf{$\model$ (Ours)} & \cellcolor{cGrey}CLIP & \cellcolor{cGrey}\textbf{4.24} & \cellcolor{cGrey}\textbf{3.99} & \cellcolor{cGrey}\textbf{16.28} & \cellcolor{cGrey}\textbf{5.80} & \cellcolor{cGrey}\textbf{3.25} & \cellcolor{cGrey}\textbf{22.39} & \cellcolor{cGrey}\textbf{21.95} & \cellcolor{cGrey}\textbf{22.17} \\
    \bottomrule[2pt]
\end{tabular}%
}
\caption{Comparison with previous methods on YouCook2 validation set. 
$\model$ achieves the best performance in both captioning and localization metrics.
BLEU@N denotes the average of BLEU-1 through BLEU-4. * denotes results reported in~\cite{sail}.
}
\label{table:yc2}
\end{table*}

\subsection{Model Training and Inference}
Following SAIL~\cite{sail}, our training objective consists of a captioning 
loss $\mathcal{L}^{\text{cap}}$ and a contrastive loss $\mathcal{L}^{\text{con}}$, 
supplemented by our proposed gated attraction loss $\mathcal{L}^{\text{attr}}$.
Specifically, $\mathcal{L}^{\text{cap}}$ trains the model to reconstruct 
the GT caption {$C_n$} from event region ${\mathbf{v} \odot M_{n,i}^{evt}}$ and its complement ${\mathbf{v} \odot (1-M_{n,i}^{evt}})$ with CE loss. 
$\mathcal{L}^{\text{con}}$ encourages the visual feature 
of each event region ${\mathbf{v} \odot M_{n,i}^{evt}}$ to be aligned with the text feature of its corresponding caption $C_n$, utilizing margin ranking loss in the CLIP~\cite{clip} feature space.
 
Unlike SAIL, $\model$ selectively applies $\mathcal{L}^{\text{attr}}$ using $g_n$ {from ~\Cref{eqn:gate}}:
\begin{equation}
    \mathcal{L}_n^{\text{attr}} = g_n \cdot \left(1 - \text{cos}(\bar{\mathbf{v}}_n^{'*},\ \mathbf{z}_{j_n})\right),
\end{equation}
where $g_n$ modulates the loss magnitude according to the transition confidence.

Additionally, we apply a similarity-based filtering step to guard against low-quality 
VLM captions that may not accurately describe the visual content. Specifically, an 
inter-event region is included in the loss computation only if the alignment score 
$s_n^*$ exceeds a predefined threshold $\theta$ 
(i.e., $s_n^* \geq \theta$).
This restricts training to reliably aligned inter-event pairs, preventing noisy or hallucinated captions from harming learning.
Let $\mathcal{A} = \left\{ n \mid g_n \geq 0.5 \ \text{and} \ s_n^* \geq \theta \right\}$ denote the set of all accepted intervals; the final attraction loss is:
\begin{equation}
    \mathcal{L}^{\text{attr}} = \frac{1}{|\mathcal{A}|} \sum_{n \in \mathcal{A}} \mathcal{L}_n^{\text{attr}}.
\end{equation}
The overall training objective combines the base loss with the attraction loss, weighted by $\lambda^{\text{attr}}$:
\begin{equation}
    \mathcal{L} = \mathcal{L}^{\text{cap}} + \mathcal{L}^{\text{con}} + \lambda^{\text{attr}} \mathcal{L}^{\text{attr}}.
\end{equation}

During inference, following~\cite{ilcacm}, the model first generates boundary-free captions to determine the event count, then produces Gaussian masks for each event to extract $(c_n, w_n)$, which are mapped to timestamps, and refines the captions using event-specific masked features.
More details are provided in the supplementary material.

\section{Experiments}
\label{subsec: setting}
\noindent\textbf{Datasets.}
We evaluate our method on two widely used DVC benchmarks.
{ActivityNet Captions}~\cite{ActivityNet} contains 20K untrimmed videos averaging 120 seconds, each annotated with approximately 3.7 temporally localized events.
{YouCook2}~\cite{youcook2} consists of around 2K untrimmed cooking videos with an average duration of 320 seconds, where each video is accompanied by 7.7 localized events on average.

\noindent\textbf{Evaluation Metrics.}
We assess performance on both captioning and localization subtasks of DVC.
For captioning quality, we report METEOR~\cite{banerjee2005meteor}, CIDEr~\cite{vedantam2015cider}, ROUGE-L~\cite{lin2004rouge}, and BLEU-N~\cite{papineni2002bleu} using the official evaluation tool~\cite{ActivityNet}, along with SODA\_c~\cite{soda}.
For event localization, we report mean Average Precision, mean Average Recall, and F1 score.
All metrics are computed across IoU thresholds of \{0.3, 0.5, 0.7, 0.9\} and averaged.

\noindent\textbf{Implementation Details.}
We adopt Distilled-GPT2~\cite{GPT2} as the caption decoder and optimize all parameters with AdamW, following~\cite{ilcacm}.
On ActivityNet Captions, the learning rate is initialized to 1e-4, and training runs for 10 epochs in both the captioning and localization stages. 
Also we use $\alpha = 0.5$, $\beta = 2$, $\theta = 0.2$, $\Omega = \{0.2, 0.4, 0.6\}$, and $\lambda_{\text{attr}} = 0.4$.
On YouCook2, we train 4 and 25 epochs for the captioning and localization stages, respectively.
All training is performed on a single NVIDIA A6000 GPU.
Additional details are provided in the supplementary material.

\begin{table}[t]
    \centering
    \small
    \renewcommand{\arraystretch}{0.95}
    \resizebox{0.48\textwidth}{!}{
        \begin{tabular}{ccc|ccc|c}
            \toprule[1.5pt]
            \makecell{\textbf{VLM}\\\textbf{Caption}} &
            \makecell{\textbf{Adaptive}\\\textbf{Gate}} &
            \makecell{\textbf{Adaptive}\\\textbf{Mask}} &
            \multicolumn{3}{c|}{\textbf{Captioning}} &
            \multicolumn{1}{c}{\textbf{Localization}} \\
             \cmidrule(lr){4-6} \cmidrule(lr){7-7}
            {\small (1)} & {\small (2)} & {\small (3)} & \textbf{S\_c} & \textbf{R-L} & \textbf{C} & \textbf{F1} \\
            \midrule
            \ding{55} & \ding{55} & \ding{55} & 6.34 & 15.26 & 35.03 & 56.86 \\
            \ding{51} & \ding{55} & \ding{55} & 6.29 & {15.45} & 35.73 & {57.73} \\
            \ding{51} & \ding{51} & \ding{55} & {6.45} & 15.43 & \underline{36.61} & {57.67} \\
            \ding{51} & \ding{55} & \ding{51} & \underline{6.48} & \underline{15.50} & {36.51} & \underline{57.80} \\
            \rowcolor{cGrey}
            \ding{51} & \ding{51} & \ding{51} & \textbf{6.49} & \textbf{15.60} & \textbf{36.87} & \textbf{58.18} \\
            \bottomrule[1.5pt]
        \end{tabular}
    }
    \caption{Ablation study on key components. We first compare {(1)} the presence of VLM and then investigate the contributions of {(2)} Narrative-Aware Inter-Event Selection and {(3)} Adaptive Inter-Event Masks components within the VLM-based framework.}
    \label{table:component_ablation}
\end{table}

\subsection{Comparison with State-of-the-Art}
\Cref{table:main} summarizes the captioning and localization results on ActivityNet Captions.
$\model$ achieves the best performance across both tasks among all weakly-supervised methods, obtaining a CIDEr of 36.87 and an F1 score of 58.18, which surpasses the previous state-of-the-art SAIL~\cite{sail}.
The localization gains are attributed to improvements in both recall (56.13) and precision (60.38), indicating that our adaptive inter-event mechanism enhances temporal boundary estimation.
Notably, $\model$ even outperforms several fully-supervised methods on the majority of metrics, despite the absence of temporal boundary annotations during training.
As shown in~\Cref{table:yc2}, these improvements generalize to YouCook2, where $\model$ again achieves the highest scores in both captioning and localization among WSDVC methods.
The consistent gains across two benchmarks validate the effectiveness of our approach.

\begin{figure*}[h]
\begin{center}
\includegraphics[width=0.95\linewidth]{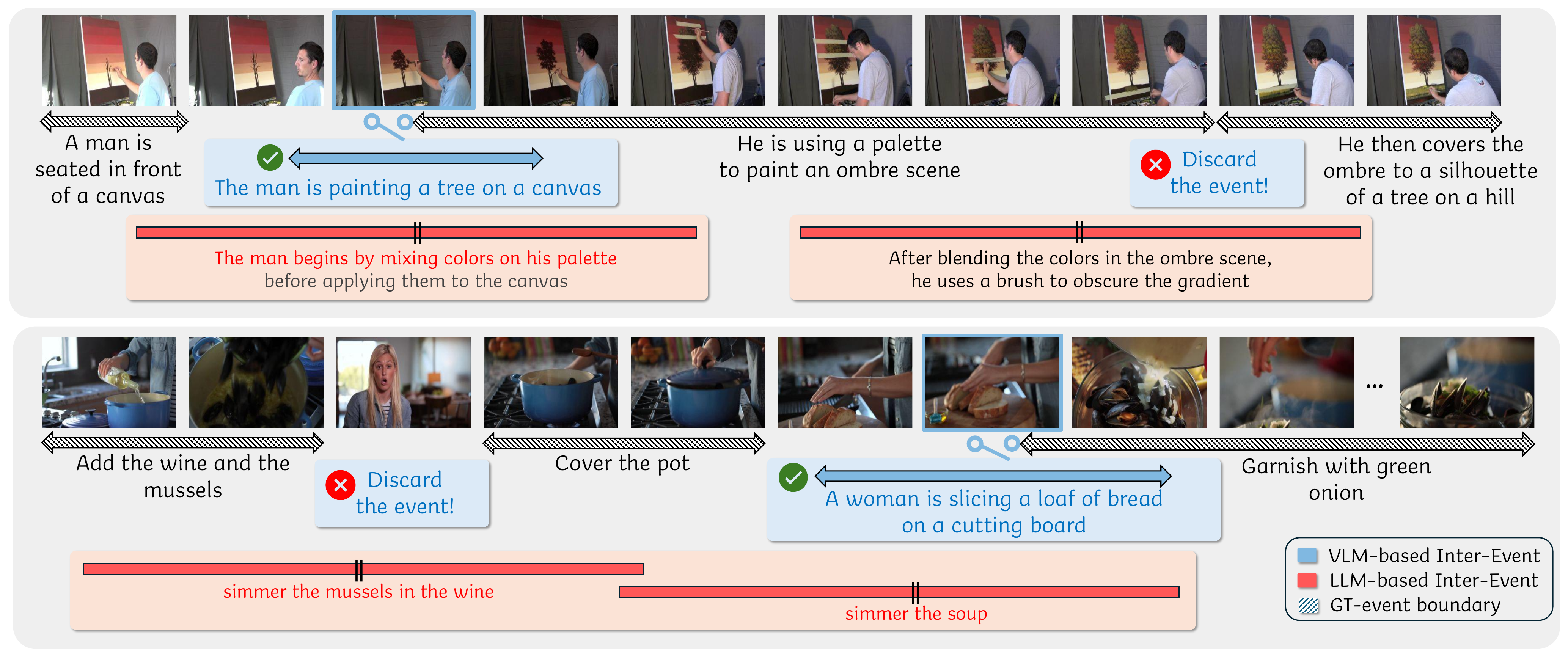}
\end{center}
\caption{Qualitative results about transition events.
Prior LLM-based method injects a transition into every gap, including uninformative ones. 
{Also,} even for meaningful transitions, they describe regions misaligned with the actual transition, resulting in inaccurate captions.
In contrast, $\model$ generates transitions only where a meaningful change occurs, producing captions that are well aligned with both the actual transition and the underlying frames.}
\label{fig:LLM_VLM_qualitative}
\end{figure*}
\subsection{Ablation Studies}

\noindent\textbf{Component Ablation.}
In~\Cref{table:component_ablation}, we analyze the contribution of each component.
Replacing LLM-synthesized captions with VLM-generated ones already improves CIDEr, ROUGE-L and F1, confirming the benefit of visually grounded inter-event descriptions.
Adding Narrative-Aware Inter-Event Selection improves almost all captioning scores, showing that selective inter-event guidance suppresses noise while retaining informative transitions.
Furthermore, Adaptive Inter-Event Masks provide further improvements, showing that tailoring each mask's center and width to the content captures transitions better than fixed templates.
Finally, combining all components achieves the best performance, confirming that auxiliary supervision tailored to each video is key to both captioning and localization.

\begin{table}[t]
    \centering
    \resizebox{0.45\textwidth}{!}{
    \begin{tabular}{c|ccc|c} %
        \toprule[2pt]
        \multirow{2}{*}[-0.5ex]{\textbf{Method}} & \multicolumn{3}{c|}{\textbf{Captioning}} & \multicolumn{1}{c}{\textbf{Localization}} \\
        \cmidrule(lr){2-4} \cmidrule(lr){5-5} 
        & \textbf{S\_c} & \textbf{R-L} & \textbf{C}  & \textbf{F1} \\
        \midrule
       SAIL (LLM) & 6.29   & 15.29   & 35.38 & 57.00 \\
        \midrule
        InternVL3-1B  &\underline{6.47}  & \underline{15.52}   & 36.71 & \underline{57.98} \\

        Qwen2.5-VL-3B & {6.43}  & {15.49}   & \underline{36.82} & {57.62} \\\

        xGen-MM-4B & \underline{6.47}  & {15.45}   & {36.45} & {57.86} \\

        SmolVLM2-2.2B & {6.44}  & {15.49}   & {36.50} & {57.94} \\

        \rowcolor{cGrey} 
        BLIP-2-2.7B& \textbf{6.49} & \textbf{15.60} & \textbf{36.87} & \textbf{58.18} \\
        \bottomrule[2pt]
    \end{tabular}
    }
    \caption{Ablation study on various VLMs, illustrating that $\model$ consistently improves performance regardless of the choice of the VLM model.}
    \label{table:VLM}
\end{table}

\noindent\textbf{Analysis of VLM Captions.}
\Cref{table:VLM} examines the impact of the captioning model used to generate frame-level descriptions.
All VLM-based variants (BLIP-2~\cite{li2023blip2}, InternVL3~\cite{zhu2025internvl3}, Qwen2.5-VL~\cite{bai2025qwen3}, xGen-MM~\cite{blip3} and SmolVLM2~\cite{smolvlm}) consistently outperform LLM-based SAIL across all metrics, confirming that visually grounded captions provide more reliable supervision than video-blind linguistic synthesis. 
These results suggest that the primary gain stems from grounding captions in actual visual content rather than from the choice of the VLM. \Cref{fig:LLM_VLM_qualitative} further illustrates this distinction qualitatively: while LLM-based captions often hallucinate events absent from the video due to their reliance on linguistic context alone, VLM-generated captions faithfully describe what appears in each frame.

\noindent\textbf{Analysis of Gate Activation Criteria.}
To examine which signal best identifies genuine inter-event transitions, we compare three criteria for opening the adaptive gate: random activation, raw CLIP visual features, and our VLM-generated captions. 
As shown in~\Cref{table:gate_activation}, random activation performs the worst, while raw visual features improve over random but still fall short, since low-level signals are easily confounded by non-semantic variation.
In contrast, our caption-based criterion achieves the best scores on both captioning and localization, showing that semantically abstract textual descriptions are a more reliable indicator of underlying transitions for adaptive gap selection.

\begin{figure}[t]
\begin{center}
\includegraphics[width=0.8\linewidth]{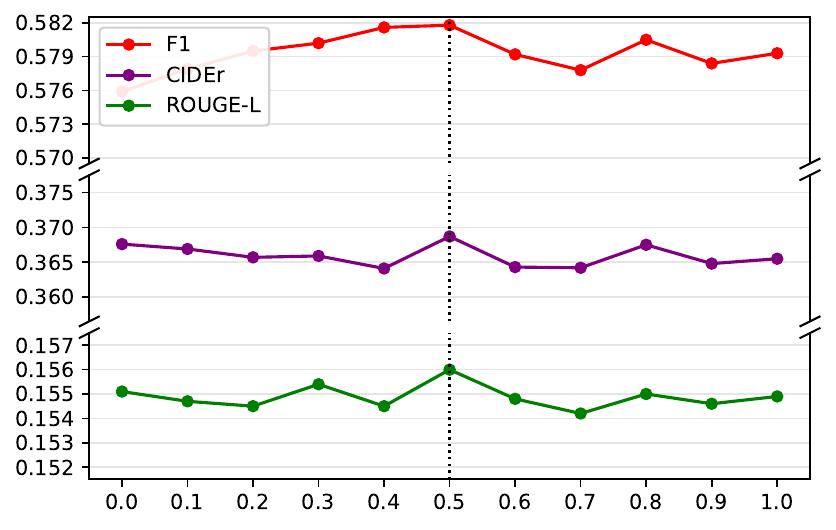}
\end{center}
\caption{Hyperparameter search for the interpolation coefficient $\alpha$. The X-axis denotes $\alpha$, and the Y-axis reports captioning and localization scores.
SAIL's fixed midpoint~\cite{sail} $(\alpha=0)$ yields the lowest scores, whereas interpolating the semantic change point $(\alpha>0)$ consistently improves performance.}
\label{fig:textpeak_alpha}
\end{figure}

\begin{table}[t]
    \tabcolsep=12pt
    \centering
    \renewcommand{\arraystretch}{0.98}
    \resizebox{0.45\textwidth}{!}{
    \begin{tabular}{c|ccc|c} %
        \toprule[2pt]
        \multirow{2}{*}[-0.5ex]{\textbf{Method}} & \multicolumn{3}{c|}{\textbf{Captioning}} & \multicolumn{1}{c}{\textbf{Localization}} \\
        \cmidrule(lr){2-4} \cmidrule(lr){5-5} 
        & \textbf{S\_c} & \textbf{R-L} & \textbf{C}  & \textbf{F1} \\
        \midrule
       SAIL & {6.29} & 15.29 & 35.38 & 57.00 \\
        \midrule

        Random & {6.26} & \underline{15.32} & {35.87} & {56.88} \\
        
        Raw Video & \underline{6.39} & \underline{15.32} & \underline{36.34} & \underline{57.42} \\

        \rowcolor{cGrey} 
        Caption & \textbf{6.49} & \textbf{15.60} & \textbf{36.87} & \textbf{58.18} \\
        \bottomrule[2pt]
    \end{tabular}
    }
    \caption{Ablation study on gate activation criteria for inter-event supervision.
    Textual semantic guides outperform raw visual features for adaptive gap selection.}
    \label{table:gate_activation}
\end{table}
\begin{table}[t]
    \tabcolsep=12pt
    \centering
    \resizebox{0.45\textwidth}{!}{
    \begin{tabular}{c|ccc|c}
        \toprule[1.5pt]
        \textbf{Method} & \textbf{S\_c} & \textbf{C} & \textbf{F1} & \textbf{Cos Sim} \\
        \midrule
        SAIL & 6.29 & 35.38 & 57.00 & 0.1460 \\
        \model \;(w/o F) & 6.43 & 36.41 & 57.89 & 0.2624 \\
        \rowcolor{cGrey} 
        \model & \textbf{6.49} & \textbf{36.87} & \textbf{58.18} & \textbf{0.2699} \\
        \bottomrule[1.5pt]
    \end{tabular}
    }
    \caption{Cosine similarity between inter-event visual features and caption features.
    Our method more faithfully represents inter-event regions.
    w/o F denotes without similarity filtering method.}
    \label{table:inter_sim}
\end{table}

\noindent\textbf{Analysis of Inter-Masks.}
As shown in \Cref{fig:textpeak_alpha}, performance consistently improves when both the temporal midpoint and the semantic transition point are jointly considered, compared with relying solely on the temporal midpoint.
In particular, intermediate values of $\alpha$  yield stable gains across all three metrics, with the best performance achieved at $\alpha=0.5$, where the model attains the highest captioning and localization scores. 
Also, \Cref{fig:LLM_VLM_qualitative} shows that the adaptive masks align well with the event transition points in the video.

\noindent\textbf{Analysis of Inter-Events.}
To verify whether our VLM captions and adaptive mask construction effectively represent inter-event regions, we measure the cosine similarity between the pooled inter-event visual features and their corresponding caption features.
As shown in~\Cref{table:inter_sim}, we observe a substantial increase in average similarity compared to the LLM-based baseline.
This confirms that our approach, which combines visually grounded VLM captions with adaptive center blending and width selection, more faithfully captures the underlying inter-event content, ultimately providing higher-quality auxiliary supervision for training.

\noindent\textbf{Comparisons with MLLM-based methods.}
We compare MLLM-based models~\cite{timechat,vtgllm,trace,timeexpert} that explicitly address temporal grounding and dense captioning (\Cref{table:R12}).
Despite fully supervised training with large-scale data and substantially larger models, these MLLM-based methods still struggle with the compound challenge of simultaneously performing grounding and captioning. 
In contrast, $\model$ effectively handles this task even under the weakly-supervised setting without any temporal annotations, while using a much smaller model.

\begin{table}[t]
    \centering
    \resizebox{0.5\textwidth}{!}{
    \begin{tabular}{c|cc|cc|c} 
        \toprule[1.5pt]
        \multirow{2}{*}[-0.5ex]{\textbf{Model}} & \textbf{Fully} & \textbf{Backbone} & \multicolumn{2}{c|}{\textbf{Captioning}} & \multicolumn{1}{c}{\textbf{Localization}} \\
        \cmidrule(lr){4-5} \cmidrule(lr){6-6} 
        & \textbf{Supervised} & \textbf{Params} & \textbf{S\_c} & \textbf{C}  & \textbf{F1} \\
        \midrule
        TimeChat & \ding{51} & 7B & 4.7 & 19.0 & 36.9 \\
        VTG-LLM & \ding{51} & 7B & 5.1 & 20.7 & 34.8 \\
        TRACE & \ding{51} & 7B & 6.0 & 25.9 & 39.3 \\
        TimeExpert & \ding{51} & 5.9B & \textbf{6.5} & 28.4 & 40.5 \\
        \rowcolor{cGrey} 
        \model & \ding{55} & 133M & 6.49 & \textbf{36.87} & \textbf{58.18} \\
        \bottomrule[1.5pt]
    \end{tabular}
    }
    \caption{Comparison with MLLM-based models on ActivityNet Captions. Despite using weak supervision and a much smaller model, $\model$ outperforms fully-supervised MLLM-based methods.}
    \label{table:R12}
\end{table}

\noindent\textbf{Direct Evaluation of the Gate on Human-Annotated Transitions.}
\begin{table}[t]
    \tabcolsep=12pt
    \centering
    \resizebox{0.4\textwidth}{!}{
    \begin{tabular}{c|cc|c}
        \toprule[1.5pt]
        \textbf{Method} &  \textbf{Recall} & \textbf{Precision} & \textbf{F1} \\
        \midrule
        SAIL  & 100 & 37.89 & 54.96 \\
       Random  & 37.90 & 49.48 & 42.92 \\
        \rowcolor{cGrey} 
        \model  & \textbf{74.99} & \textbf{64.29} & \textbf{69.23} \\
        \bottomrule[1.5pt]
    \end{tabular}
    }
    \caption{Comparison of transition identification performance on human-annotated gaps.
    $\model$'s gate substantially outperforms both always-inject (SAIL) and random gating in F1.}
    \label{table:gate_eval}
\end{table}
Since no ground-truth annotations exist for inter-event transitions, we constructed a manually verified transition validation set, then directly evaluated our gate on it.
We designed a multi-stage protocol, with human verification as the final arbiter.
(1) Gap extraction and sampling. From consecutive GT event pairs in ActivityNet validation videos, we randomly sampled 200 inter-event gaps.
(2) Initial transition proposal. Frames from each gap (1 fps) were fed into Gemini 3.5 Flash, prompted to judge whether a distinct transitional event—not covered by either adjacent GT caption—occurs in the gap, explicitly excluding camera cuts, angle changes, and mere continuations.
(3) Cross-verification. Each gap was independently re-judged by GPT-5.5 Pro with the identical prompt, without revealing the first model's answer. Agreed labels became draft labels; disagreements were excluded.
(4) Human verification. All draft labels were then verified by human annotators, blind to our model's outputs. We retained only gaps with unanimous agreement, to ensure label reliability. This yielded a final validation set of 95 gaps (36 gaps with a transition, 59 gaps without).
On this set, we evaluated the gate decision for each annotated gap (\Cref{table:gate_eval}).
Our gate mechanism achieves substantially better transition detection than both SAIL's always-inject strategy and random gating, suggesting that text-level semantic change is an effective signal for identifying genuine transitions and directly substantiating the whether-to-inject component of our central claim.

\noindent\textbf{Computational Cost Analysis.}
\begin{table}[t]
    \centering
    \resizebox{0.48\textwidth}{!}{ 
        \begin{tabular}{c|ccc}
        \toprule[1.5pt]
            \textbf{Method} & \textbf{Train time} &
            \textbf{Inference time}&
            \textbf{GPU usage} 
            
            \\
            \midrule

            ILCACM & {1H 42M 31S} & {7M 16S} &{33.08 \text{GiB}}   

            \\

             SAIL & {1H 49M 50S} & {7M 35S} &{33.11 \text{GiB}}   

            \\
            \rowcolor{cGrey} 
            $\model$  &{1H 52M 53S} & {7M 51S} &{33.13 \text{GiB}}  
            \\

            \bottomrule[1.5pt]
        \end{tabular}
    }
    \caption{Model training computational cost comparison. The additional cost of $\model$ is negligible across all three metrics.}
    \label{table:appendix_computational}
\end{table}
We measure the time and memory consumed during training and inference, averaged over five runs.
As shown in~\Cref{table:appendix_computational}, the training and inference times of our method are nearly identical to those of the baseline~\cite{ilcacm}, and the memory consumption is comparable as well. 
Since captions are extracted in advance and only their features are used, the captioning process adds almost no overhead, and the mask width selection  is a simple dot product with negligible effect on the overall training time.


\section{Conclusion}
We present $\model$, a framework that rethinks how inter-event information is utilized in WSDVC.
Our key insight is twofold: (1) transitions should be applied selectively, only where the video exhibits a transition, rather than uniformly across all gaps; (2) each transition should be localized according to the actual video content, rather than fixed at a heuristic midpoint. 
To this end, $\model$ repurposes a VLM as a transition-search tool, grounding both decisions—\textit{whether} and \textit{where}—in visual evidence. 
Experiments on ActivityNet Captions and YouCook2 confirm that $\model$ sets a new state-of-the-art in both captioning and localization, demonstrating the importance of adaptive, visually grounded inter-event utilization for WSDVC.

\section*{Limitations}
Our method grounds transition discovery in VLM-generated frame captions, so its effectiveness is bounded by the quality of those captions: when the VLM produces generic, repetitive, or inaccurate descriptions—particularly in domains underrepresented in its pre-training—the caption-space dissimilarity signal becomes unreliable, causing the adaptive gate to miss genuine transitions or open on spurious ones. 

\section*{Acknowledgments}
This work was partly supported by the Institute of Information \& Communications Technology Planning \& Evaluation (IITP) grant funded by the Korean government (MSIT) RS-2025-25422680, Metacognitive AGI Framework and its Applications and the AI Seoul Tech Research Support Program of the Seoul Future Foundation.


\bibliography{custom}

 \appendix

 \clearpage
\section*{Appendix}
\label{sec:appendix}

\setcounter{page}{1}
\setcounter{table}{0}
\setcounter{figure}{0}
\renewcommand{\thetable}{A.\arabic{table}}
\renewcommand{\thefigure}{A.\arabic{figure}}

In this Appendix, we provide additional details and qualitative results to support our findings. 
\section{Additional Implementation Details}
\noindent{\textbf{Frame Sampling.}}
Following ILCACM~\cite{ilcacm} and SAIL~\cite{sail}, we uniformly sample 32 and 100 frames per video for ActivityNet Captions and YouCook2, respectively.

\noindent{\textbf{VLM Settings.}}
For caption generation with BLIP-2~\cite{li2023blip2}, we employ the {blip2-opt-2.7b} model and use the prompt \textit{``What is happening in this image?''}. 
For InternVL3 and Qwen2.5-VL, we use the prompt \textit{``Describe this image briefly in one sentence.''}, with the maximum number of newly generated tokens set to 30.

\noindent{\textbf{Additional Details.}}
We set the number of event queries to 22 for ActivityNet Captions and 18 for YouCook2. For the video encoder, we follow the ILCACM~\cite{ilcacm} architecture, which consists of a single Conv1D layer (kernel size 5) followed by a single Transformer decoder layer. To prevent overfitting, we additionally apply label smoothing during training.

For inter-event processing, we sort the predicted event masks by their temporal centers such that $c_1 \leq c_2 \leq \cdots \leq c_{N_e}$, following~\cite{ilcacm}. 
Since gaps spanning only a few frames are unlikely to yield meaningful transitions or reliable statistics, we apply Narrative-Aware Inter-Event Selection only to gaps containing at least 4 frames, discarding any shorter gap.

\noindent{\textbf{Inference Details.}}
At test time, our model follows the same inference procedure as~\cite{ilcacm} to generate temporally localized event captions. 
First, the complete video embedding is decoded with a global context prompt, i.e., ``[FULL]'', producing an initial set of event descriptions while adaptively estimating the number of events in the video. 
The mask generation module then predicts the temporal center and width of each identified event, from which Gaussian attention masks are constructed. 
Finally, each masked event representation is decoded with event-specific conditioning, i.e., ``[MASK] 1 events:'', to refine the initial captions and generate more precise event descriptions.

Importantly, the VLM-generated narratives in $\model$ are generated offline before training and are used only to construct selective inter-event supervision during training. 
Therefore, $\model$ introduces no VLM dependency at inference time. 
The trained DVC model predicts event masks and captions using the same base WSDVC pipeline.

\begin{table}[h]
    \centering
    \resizebox{0.45\textwidth}{!}{
    \begin{tabular}{c|cccc|c} %
        \toprule[2pt]
        \multirow{2}{*}[-0.5ex]{\textbf{Method}} & \multicolumn{4}{c|}{\textbf{Score}} & \multicolumn{1}{c}{\textbf{Time}} \\
        \cmidrule(lr){2-5} \cmidrule(lr){6-6}
        & \textbf{S\_c} & \textbf{R-L} & \textbf{C} & \textbf{F1} & \textbf{H} \\
        \midrule
        SAIL{$^\dagger$} (LLM) & {6.29} & {15.29} & 35.38 & 57.00 & {1H 38M} \\
        $\model$ (VLM) & \textbf{6.49} & \textbf{15.60} & \textbf{36.87} & \textbf{58.18} & {1H 46M} \\
        \bottomrule[2pt]
    \end{tabular}
    }
    \caption{Offline caption-generation time. $\model$'s VLM-based generation requires comparable offline generation time to SAIL's LLM-based synthesis (1H 46M vs. 1H 38M), while yielding better captioning and localization performance. 
    $\dagger$ denotes that SAIL's generation time was measured under our own re-implementation.}
    \label{table:appendix_time}
\end{table}

\section{Computational Cost}
We measure two types of cost: the offline caption generation time and the training/inference overhead.
For caption generation, we follow the LLM inference procedure described in SAIL~\cite{sail} to synthesize transition captions on ActivityNet Captions.
As shown in~\Cref{table:appendix_time}, our VLM-based caption generation requires comparable offline time to SAIL's LLM-based synthesis (1H 46M vs. 1H 38M), while yielding clearly better captioning and localization scores.
Importantly, captions are generated offline before training, and only their precomputed features are loaded during training, so this step incurs no cost in the training loop.

\section{More Qualitative Results}
\begin{figure*}[h]
\begin{center}
\includegraphics[width=0.95\linewidth]{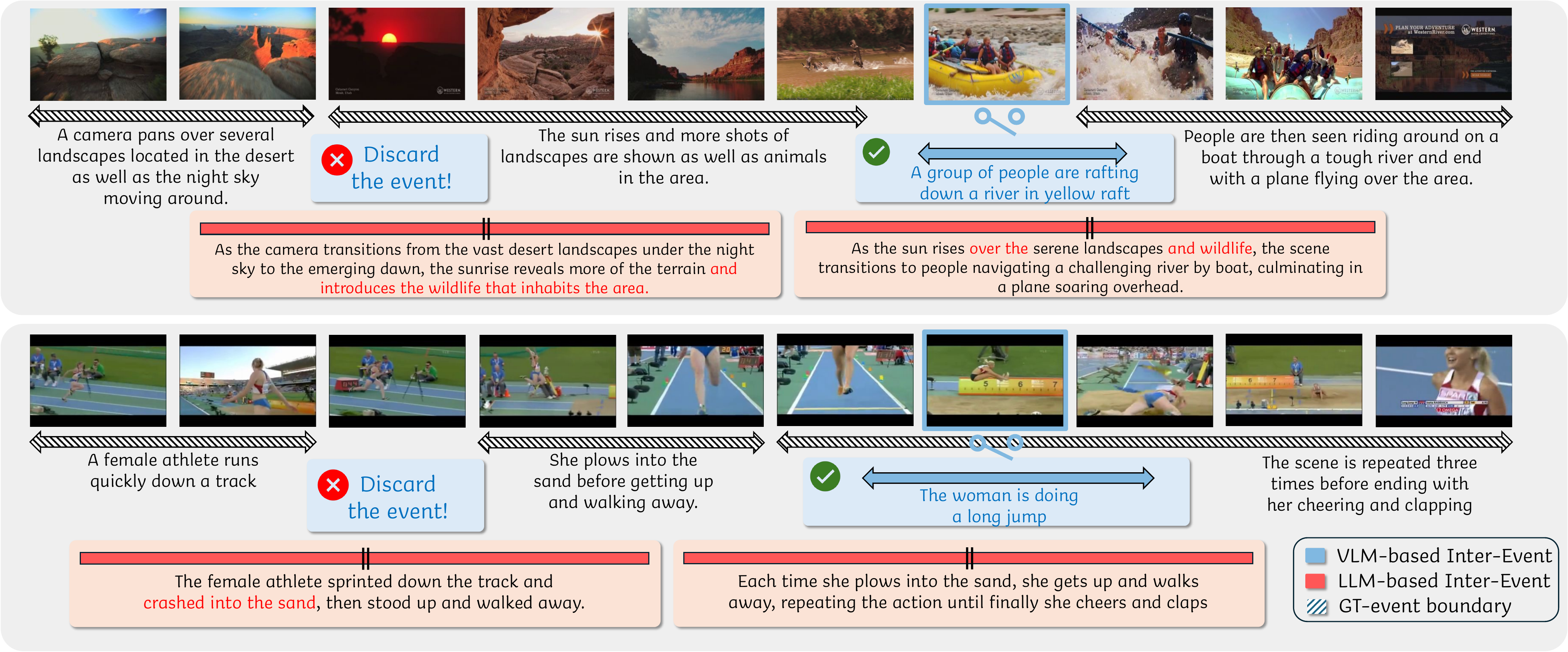}
\end{center}
\caption{Qualitative results about transition events on ActivityNet Captions.
}
\label{appendix_suppl_qual_anet}
\end{figure*}

\begin{figure*}[h]
\begin{center}
\includegraphics[width=0.95\linewidth]{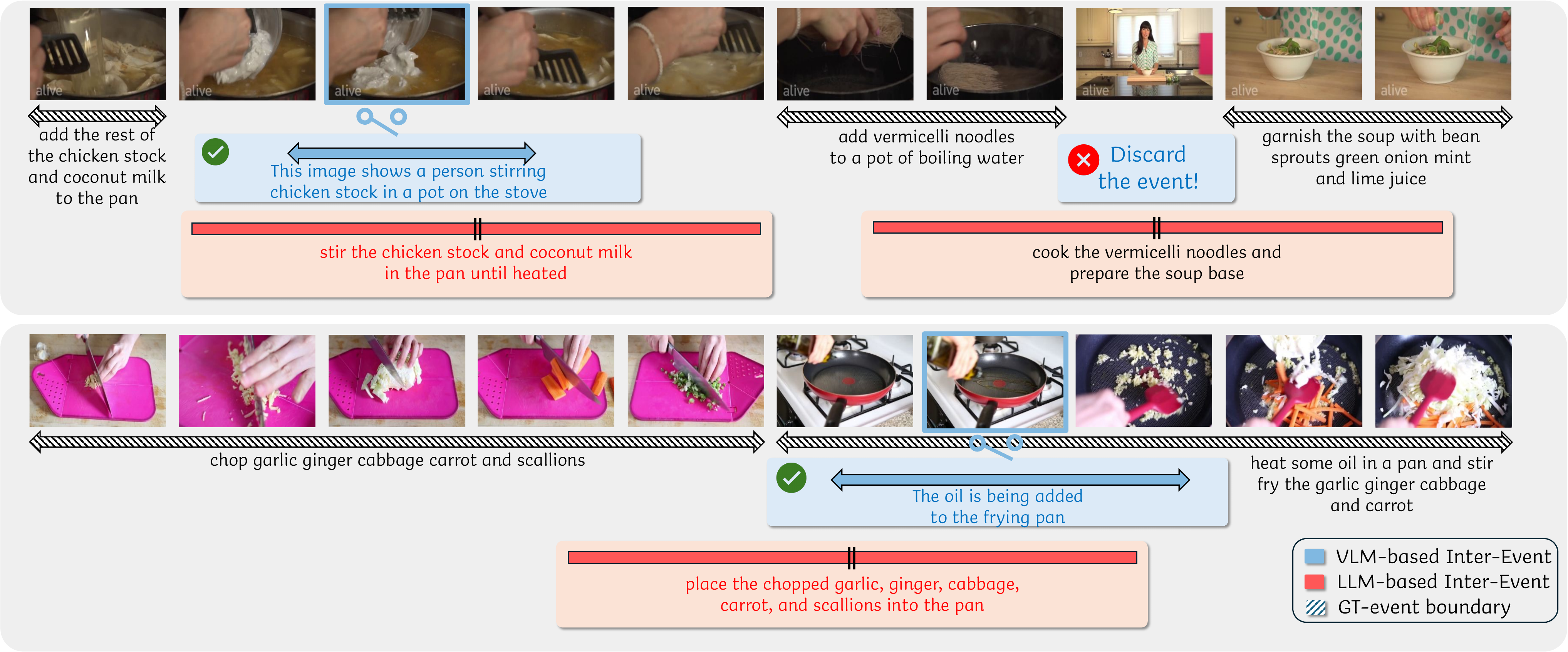}
\end{center}
\caption{Qualitative results about transition events on YouCook2.}
\label{appendix_suppl_qual_yc2}
\end{figure*}

\Cref{appendix_suppl_qual_anet} and~\Cref{appendix_suppl_qual_yc2} provide further qualitative comparisons on ActivityNet Captions and YouCook2, respectively, where blue boxes denote our VLM-based inter-event captions and red boxes denote the LLM-based ones used by prior work. 
$\model$ selectively discards uninformative gaps in which no meaningful transition occurs between adjacent events, and instead supplies a transition guide only at moments of genuine change (e.g., when rafting begins or a jump is performed).
Furthermore, whereas the LLM-based method aligns its captions with mismatched regions due to hallucinated content and a fixed placement rule, $\model$ anchors each transition at the correct temporal location, yielding captions that are faithfully aligned with the underlying frames.

\end{document}